\documentclass[acmsmall]{acmart}

\AtBeginDocument{%
  \providecommand\BibTeX{{%
    \normalfont B\kern-0.5em{\scshape i\kern-0.25em b}\kern-0.8em\TeX}}}

\begin{document}

\title{Face to Purchase: Predicting Consumer Choices with Structured Facial and Behavioral Traits Embedding}

\author{Zhe~Liu}
\affiliation{\institution{University of New South Wales}}
\email{zhe.liu@student.unsw.edu.au}

\author{Xianzhi~Wang}
\affiliation{\institution{University of Technology Sydney}}
\email{xianzhi.wang@uts.edu.au}

\author{Lina~Yao}
\affiliation{\institution{University of New South Wales}}
\email{lina.yao@unsw.edu.au}

\author{Jake~An}
\affiliation{\institution{Raiz Invest Limited}}
\email{jake@raizinvest.com.au}

\author{Lei~Bai}
\affiliation{\institution{University of New South Wales}}
\email{lei.bai@student.unsw.edu.au}

\author{Ee-Peng~Lim}
\affiliation{\institution{Singapore Management University}}
\email{eplim@smu.edu.sg}

\renewcommand{\shortauthors}{Liu, et al.}

\begin{abstract}
Predicting consumers' purchasing behaviors is critical for targeted advertisement and sales promotion in e-commerce. Human faces are an invaluable source of information for gaining insights into consumer personality and behavioral traits. However, consumer's faces are largely unexplored in previous research, and the existing face-related studies focus on high-level features such as personality traits while neglecting the business significance of learning from facial data. We propose to predict consumers' purchases based on their facial features and purchasing histories. We design a semi-supervised model based on a hierarchical embedding network to extract high-level features of consumers and to predict the top-$N$ purchase destinations of a consumer. Our experimental results on a real-world dataset demonstrate the positive effect of incorporating facial information in predicting consumers' purchasing behaviors.
\end{abstract}



\maketitle

\section{Introduction}\label{sec:introduction}
Consumer spending accounts for a significant fraction of economic activities worldwide~\cite{kooti2016portrait}.
The past years have seen tremendous digital traces of consumer purchases in both offline and online contexts.
A better understanding of consumers' purchasing behaviors helps companies target the right consumers as well as make the appropriate promotion strategies.
Traditional methods predict consumer behaviors solely based on consumers' purchasing histories~\cite{qiu2015predicting}. They typically learn consumers' preferences over products and infer consumers' motivations to build a candidate product collection; then, they could associate the product features and predict a consumer's future purchases from the candidate products~\cite{liu2017does}, based on the learned preferences.
In this sense, consumers' behavioral traits have a significant influence on their purchasing behaviors, and multiple factors may impact this influence~\cite{liu2017does}.

Besides, facial information is an important yet largely neglected source of information for consumer behavior prediction.
Indeed, facial information opens a new door to predicting consumer behaviors as it contains a new dimension of information and is often the first source of information that people capture when meeting each other.
For example, the first glimpse of a person's face can reveal much information: gender, age, emotional state, and ethnicity; then, people can infer the person's personality, financial background, and manners merely from its appearance; eventually, they could draw higher-level conclusions such as \textit{the person is young, energetic, and may like sports}.
Such high-level conclusions are especially important in a sales context---you had better suggest a pair of sports shoes rather than casual shoes to someone who likes sports.
Previous research has demonstrated that face images contain rich information that can be extracted using computer vision technologies automatically~\cite{joo2015automated,rothe2018deep,agustsson2017apparent}.
The extracted information (such as the age~\cite{perkins1993zeroing}, gender~\cite{neff2002crest}, family structure~\cite{wilkes1995household}, social stratum~\cite{coleman1983continuing}, and nation~\cite{rich1999region}) has a great impact on consumers' shopping behaviors, making the analysis of consumers' face images to predict their purchasing decisions feasible.

However, although the effectiveness of learning people's personality traits from face images has been validated in various scenarios, e.g., predicting election outcomes~\cite{joo2015automated}, social relationship~\cite{zhang2015learning}, and job interview results~\cite{escalante2018explaining}, little research has examined the role of facial images in predicting consumers' purchasing behaviors.
Inspired by the link between face images to purchasing behaviors, in this paper, we aim to predict consumers' destinations of purchases (i.e., companies) based on their face images and purchasing histories, focusing on examining whether consumers' face is indicative of their purchasing behaviors.
This problem is at a higher level than predicting specific items and requires considering repeated purchases of consumers from the same companies.
It is critical in many scenarios, such as predicting to what extent a person would be a customer of a company instead of its competitors.

The rationale and significance of predicting consumers' purchasing destinations are also testified by the business theory~\cite{solomon2014consumer}, which indicates that companies generally pay attention to building an iconic symbol in consumers' memory.
According to this theory, the first impression that comes to the mind of a consumer to KFC might be \textit{fast food} or \textit{cheap meals}. Consequently, when the consumer decides to eat fast food, it will link this intention immediately to KFC.
Given the significance of iconic symbols in consumers' memory, it is important to analyze different sources of clues to consumer's establishment of their iconic symbols towards business from a B2C perspective.

Our proposed approach applies data-driven features and graph-based models. Specifically, we regard the similarity between faces as edges and transform the cluster information into the similarity between nodes.
Then, we apply the CNN structure in our model as the additional information to find the critical edges and to propagate the labels, given its ability to extract the hidden information from images~\cite{zhang2015learning}.
In a nutshell, we make the following contributions in this paper:
\begin{itemize}
    \item We design multi-faceted features to characterize consumers' faces and their purchasing behaviors by exploiting consumer's shopping capacity and compositions reflected by their past purchases (Section 2) and face images (Section \ref{featureExtract}).
    \item We propose a semi-supervised hierarchical model that incorporates multi-aspect face embedding using Graph Convolutional Network (GCN) and Convolutional Neural Network (CNN) to predict the consumer purchase features and purchasing choices (Section \ref{2}-\ref{sec:Label and Prediction}).
    \item We demonstrate the effectiveness of facial information, graph generation, and the select layer using experiments, case examples, and correlation analysis results (Section \ref{sec:exp}).
\end{itemize}

\section{Purchase Behavior Characterization}\label{sec:features}
We investigate a set of consumers' transaction records and face images from a local financial services company to showcase its characteristics for feature designation.
Specifically, we study a dataset that consists of 243,118 transaction records of 1,485 consumers with 2,672 companies during 1/10/2017--30/10/2017, as well as the consumers' face images. 
The transactions fall into 43 different categories and belong to two payment methods: debit and credit.
Each record consists of \textit{consumer id}, \textit{transaction date}, \textit{expense}, \textit{transaction type}, \textit{transaction category}, \textit{description}, and \textit{payee}.
Table~\ref{tab:data set Description} shows some statistics of the dataset.
\begin{table}
  \caption{Dataset Statistics}
  \label{tab:data set Description}
  \center
  \begin{tabular}{p{3cm}p{3cm}}
  \toprule
  Attributes & Details \\
  \midrule
 \#Records & 243,118 \\
\#Consumer & 1,485 \\
 \#Company & 2,672\\
\# Category & 43\\
 Time Span & 1/10$-$30/10/2017\\
 Type & Debit/Credit\\
 Amount & 0.01$-$564,245 \\
 AVG Amount & 244 \\
 \#Trans & 101$-$1,281 \\
 AVG \#Trans & 164 \\
 Total Expense & 2,680$-$3,905,577\\
 AVG Expense & 40,020 \\
 \bottomrule
  \end{tabular}
\end{table}

\subsection{Analysis of Data Characteristics}\label{sec:Data Character}
We analyze the dataset from three perspectives to learn about consumers' behavior features. 

\subsubsection{Gaps in Consumers' Expenses.}
We sort the consumers in ascending order of their total expenses to study their group features and divide consumers evenly into ten groups (from group 1 to group 10) to reduce the adverse effect of the large gap between individuals. As a result, the first group (group 1) includes the lowest-expense consumers, and the last group (group 10) has highest-expense users, and the consumers in each group spend no more than the consumers in any subsequent group.
Fig.~1 shows the total expense grows fast, and the increasing rate first drops and then soars from group 1 to group 9. We have excluded showing group 10 to see the phenomenon more evidently.
\begin{figure}
\center
\includegraphics[scale=0.4]{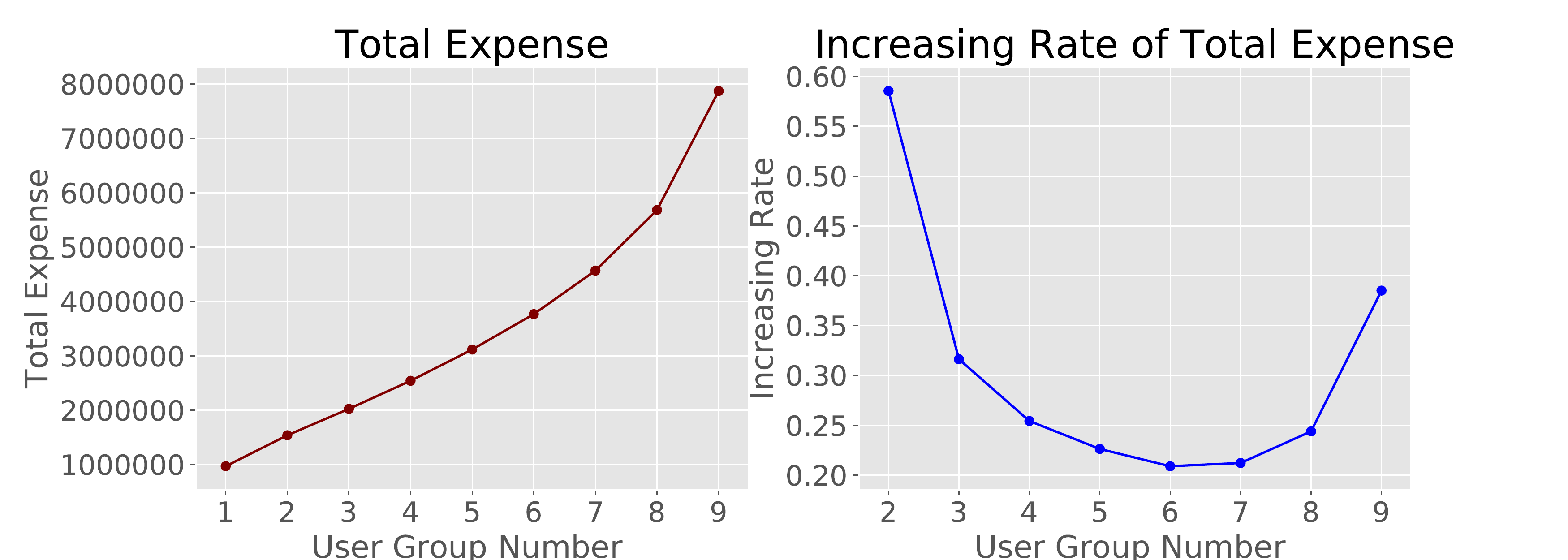}
\label{fig:highestamount}
\caption{(a): Expense of user groups. (b): Increase rate of expense.}
\end{figure}
In particular, each of the first nine groups has 149 consumers while the last group has 144 consumers.
However, the last group's expense takes up around 46\% of the total expense of all groups.
The gap of two consumers' total expenses can be as large as over \$3,900,000 (see Table~\ref{tab:data set Description}).

\subsubsection{Overlaps of Pairwise purchasing choices.}
Intuitively, consumers with similar consuming capacities should have close shopping behaviors and routines. Therefore, we investigate the relationship between consumers' shopping capacity and shopping behaviors by calculating the overlap rates of the shopping scopes of two groups:
\begin{equation}
Overlap(a,b) = \frac{\#Intersection(a,b)}{\#Union(a,b)}
\end{equation}
where $a$ and $b$ represent two arbitrary groups; \textit{\#intersection} and \textit{\#union} denote the number of the intersection and the union of the shopping company sets, respectively.
Fig.~2 shows the overlap rates of four representative groups with all the other groups, respectively, four sub-figures.
Generally, for a group with a small group number (e.g., group 1 or group 3), its overlap rate decreases as the group number of the other group increases; conversely, for a group with a large group number (e.g., group 7 or group 10), its overlap rate increases as the group number of the other group increases.
Based on the above, we conjecture three types of consumers with following characteristics:
\begin{itemize}
    \item \textit{Weak capacity consumers}: The first few groups only share more similar purchasing choices with the consumers who have low expenses in the month.
    \item \textit{Common capacity consumers}: The middle few groups share a relatively higher ratio of choices with all groups than weak and strong groups; they have more common with near groups.
    \item \textit{Strong capacity consumers}: The last few groups share more common shopping behavior with those who spend much money.
\end{itemize}
\begin{figure}
\center
\includegraphics[width=0.8\linewidth]{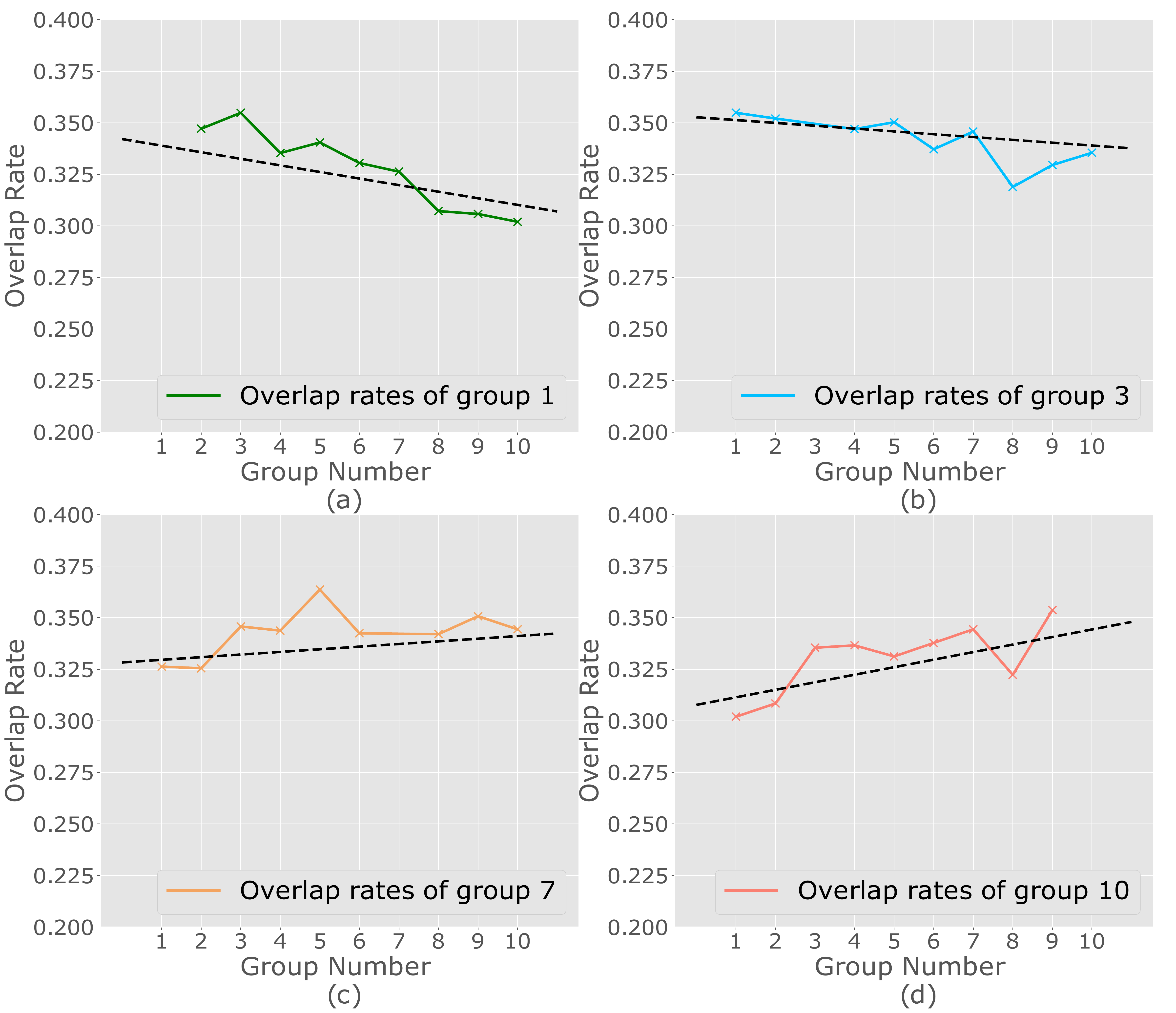}
\label{fig:overlap}
\caption{Pairwise overlap rates of four representative groups (group 1, 3, 7, and 10) with other groups.}
\end{figure}

\subsubsection{Ratio of Triple-wise Overlaps.}
We further investigate consumers' purchasing choices after analyzing the overlaps of groups. Specifically, we define \textit{impact score} to detect the shared purchase choices of two groups (e.g., group $a$ and group $b$) by calculating their triple-wise overlap rates:
\begin{equation}\footnotesize
impact(b|a) =
 \sum_{j<>a,b}^{}|\frac{\#Intersection(a,b,j)}{\#Union(b,j)}-\frac{\#Intersection(b,j)}{\#Union(b,j)}|
\end{equation}

By summing up the difference of the overlap rate of group $b$ with other groups (except $a$ and $b$) under the condition of the group $a$ and the normal one, we can know the size of the stable overlap part of $a$ and $b$.
A smaller impact score indicates a larger stable overlap; conversely, a larger overlap part means more common choices in the shopping choices of groups.

Fig.~3(a) plots the impact scores of three representative groups (group 1, group 6, group 10), while Fig.~3(b) shows the item scopes of all the groups.
The results show the user groups with wider shopping ranges and stronger shopping capacities have lower impact scores---while the first group has the smallest impact score and the last groups have the largest impact scores, the groups in the middle have the most stable overlap curves.
The above results align with our intuition that the consumers who spend more than others have a larger portfolio of purchases and share more purchasing choices with others. More specifically, wealthy consumers will be more likely to buy expensive things, while ordinary consumers can only afford the standard products. Meanwhile, the wealthy consumers purchase the standard products the same as ordinary consumers, so their diverse shopping structure leads to the observation in the triple-wise overlap rates.
%
\begin{figure}
\includegraphics[scale=0.55]{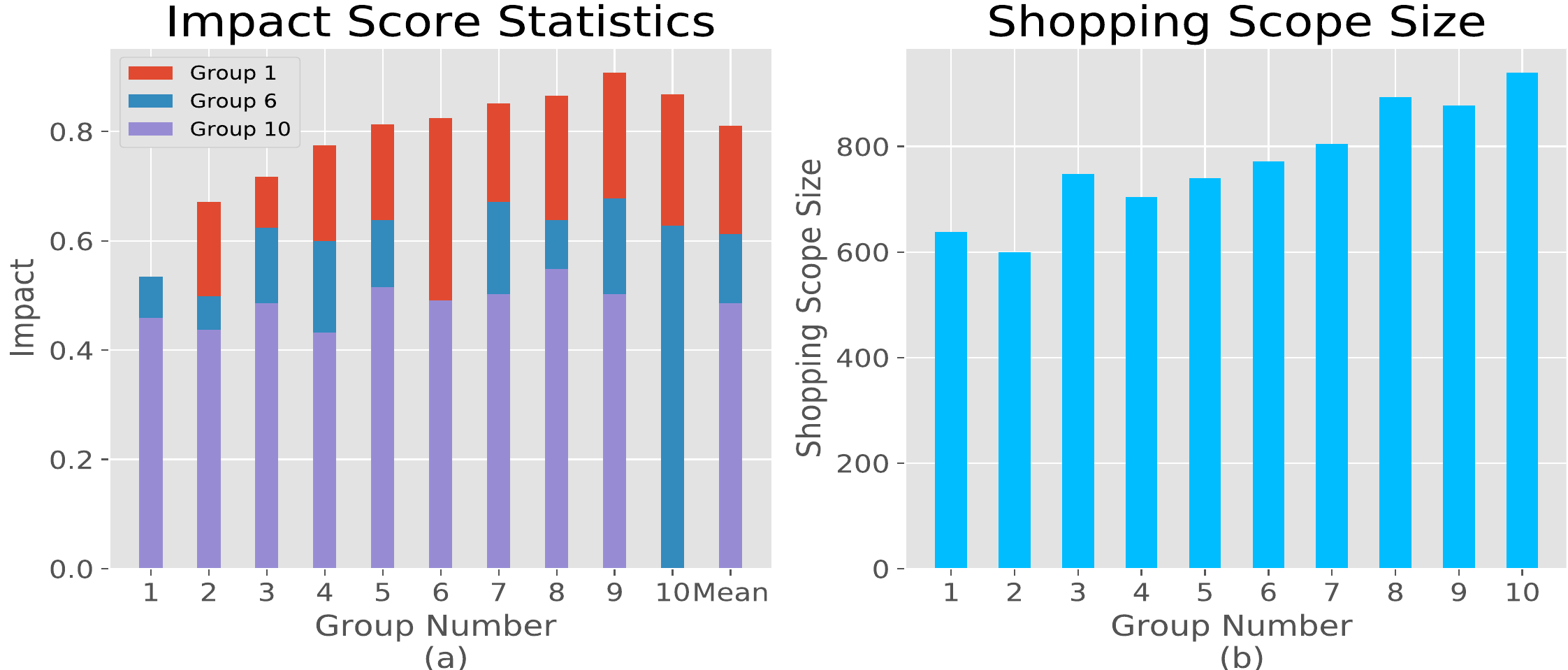}
\label{fig:impact}
\caption{(a) Impact scores of groups 1, 6, and 10.
The 'Mean' in the x-axis is the average impact of one group to other groups. (b) The shopping scope sizes of group 1 to group 9.}
\end{figure}
Since expense types remain unknown, we will discuss the shopping structures in the following sections.

\subsection{Behavioral Feature Designation}\label{sec:features}

We exploit consumers' purchasing choices in different categories of expense by analyzing purchase behaviors in Section \ref{sec:Data Character}.
Generally, consumers' purchasing behaviors have three features:
1) They are not affected by the large gaps between individuals;
2) They imply the different types of consumers and the expense;
3) They imply the shopping structures of consumers.
Accordingly, we propose two categories of consumer purchase features~\cite{liu2020you}, namely \textit{stratum features} and \textit{life features}, to portray the consumer shopping capacity and tendency, respectively.

\subsubsection{Stratum Features.}
\begin{figure*}
\center
\includegraphics[width=\textwidth,height=55mm]{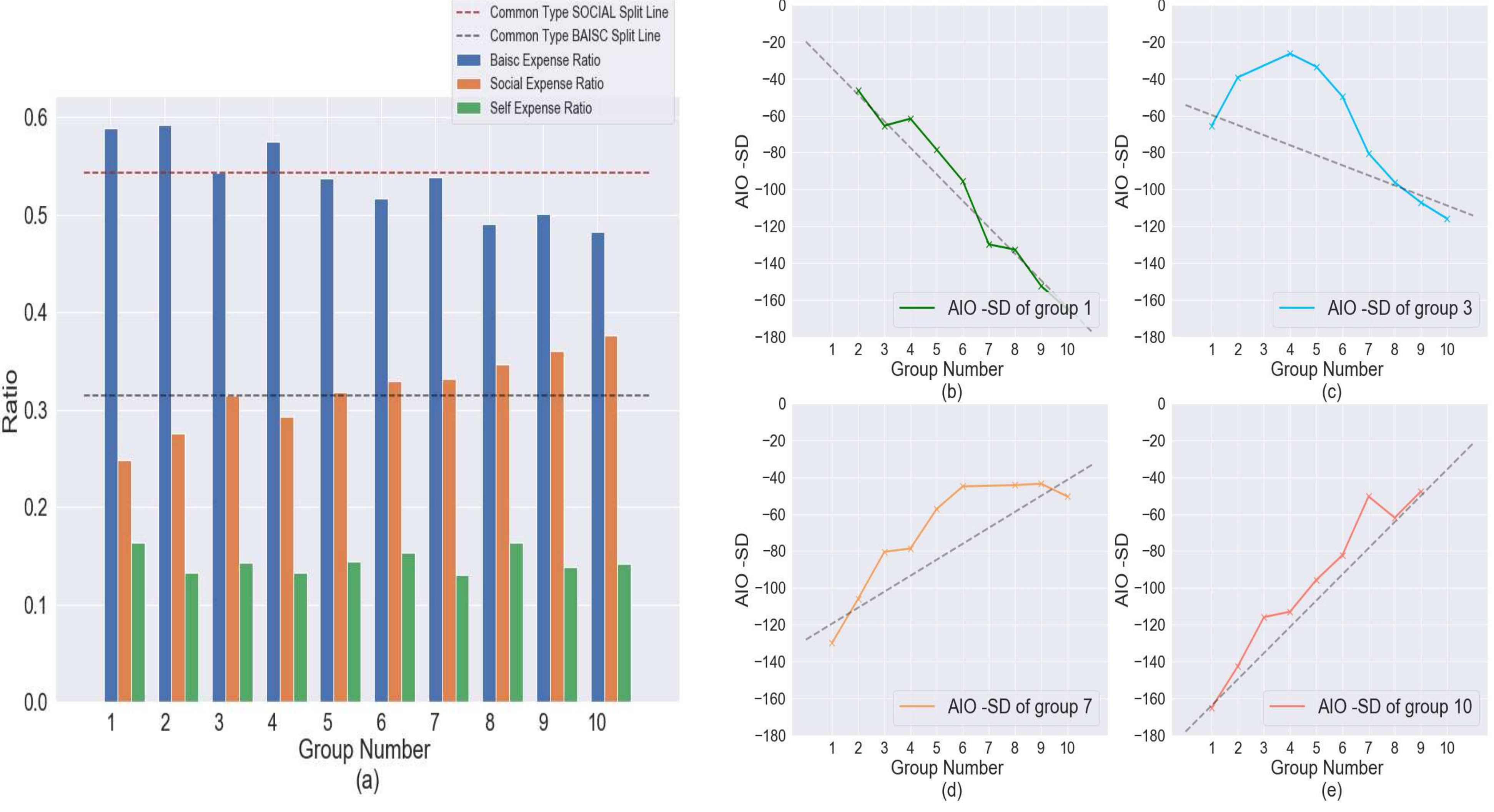}
\caption{(a) Ratio scores of stratum features of groups. The brown and black dashed lines are references to help identify different types of groups. (b)-(e): Negative AIO Standard Deviation of four representative groups (group 1, 3, 7, and 10) with other groups.}
\label{stratumX}
\end{figure*}

We design Stratum Features by dividing consumers' consumption behaviors based on Maslow's hierarchy of needs, where a higher-layer needs only occur when the lower-layer needs are satisfied, and a higher-stratum consumer normally has a higher ratio in the higher layers.
We analyze the three categories of users with differed purchasing capacities, namely weak (groups 1-2), ordinary (groups 3-7), and strong (group 8-10) by using two reference lines (the brown and black dashed lines) and considering the adjacency of groups in analyzing the scores of user groups in the dataset (shown in Fig.~\ref{stratumX}(a)).
Groups 1-2 have a weak capacity because they have higher \textit{Basic} scores above the brown bar and lower \textit{Social} scores below the black bar; in comparison, groups 8-10 demonstrate the opposite patterns and therefore have a strong capacity; lastly, we consider all the other groups (groups 3-7) to have an ordinary capacity. The three user categories have larger expense allocations on the higher-level needs from weak to strong purchasing capacity.

The original Maslow's hierarchy of needs describes five layers top the bottom: self-actualization, esteem needs, belonging and love needs, safety needs, and physiological needs.
Since some of the levels do not have clear differences, we combine the levels to three main stratum features.
Specifically, we get three categories with reference to the Maslow's hierarchy of needs~\cite{mcleod2007maslow}:
\begin{itemize}
    \item \textit{Basic}: The necessary consumption to address the fundamental physiological and safety needs such as food and water;
    \item \textit{Social}: The consumption combining esteem, belonging, and love needs for enjoyment and improved life quality;
    \item \textit{Self}: The consumption for self-actualization or mental enrichment such as education.
\end{itemize}
Our stratum features design has two advantages:
1) The Maslow's hierarchy concentrates on the ratios of each layer of consumption and can, therefore, prevent the bad effects of the large gaps between individuals;
2) Hierarchy structure: It helps classify consumers because high layers need a stronger consumption capacity.
Using this structure can exactly portray the financial expense structure, which can indicate the consumption capacity. 

\subsubsection{Life Features.}
We define life features of consumers by analyzing their shopping structure based on the AIO (Activities, Interests, and Opinions) theory~\cite{boote1980psychographics}, which cover all the life aspects of purchases.
Specifically, we look into consumers' shopping frequency and choices of targets beyond the three general types. One example is that, generally, a small portion of consumers contribute to the majority of sales in a market (the ``20:80" rule~\cite{plummer1974concept}). Another example is that a company selling clothes will be more interested in consumers who frequently buy fashion products but not in consumers who only spend money at home.

We summarize consumers' shopping frequency and choices of targets in 17 categories: \textit{Restaurant}, \textit{Entertainment}, \textit{Service}, \textit{Travel}, \textit{Shop}, \textit{Health}, \textit{Work}, \textit{Credit}, \textit{Home}, \textit{Daily}, \textit{Investment}, \textit{Bill}, \textit{Gambling}, \textit{Education},  \textit{Charity}, \textit{Fashion}, \textit{Tax}, and three stratum feature categories: \textit{Basic}, \textit{Social}, \textit{Self} (we will introduce the labeling methods in Section \ref{sec:Label and Prediction}).
To make the AIOs scores more understandable, we compress the frequency of each aspect to 0-5 relative to the maximum frequency of each aspect in the whole dataset. For each aspect, a high score indicates a more frequent buyer.
\begin{figure}
\center
\includegraphics[width=0.8\linewidth]{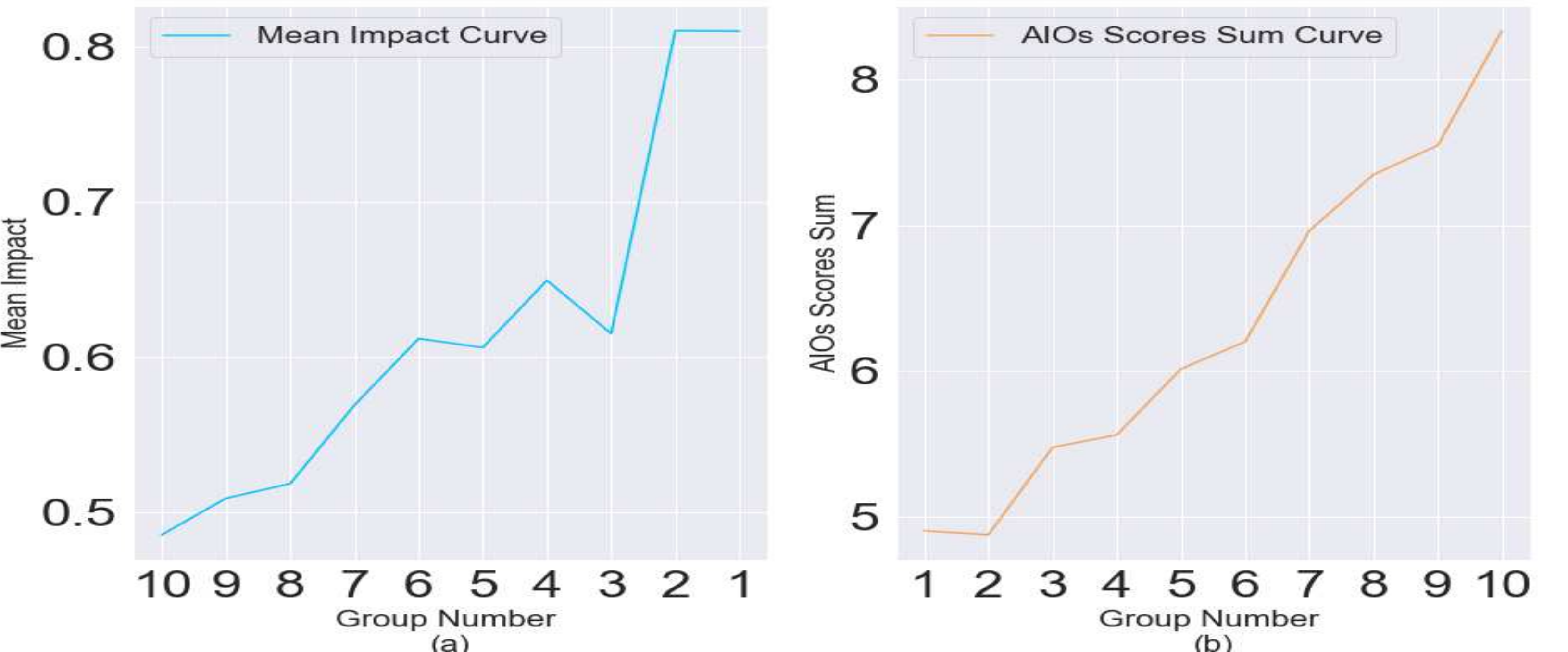}
\caption{(a) Mean impact scores. (b) AIOs scores.}
\label{tab:lifeprove}
\end{figure}
\begin{figure}
\center
\includegraphics[scale=0.4]{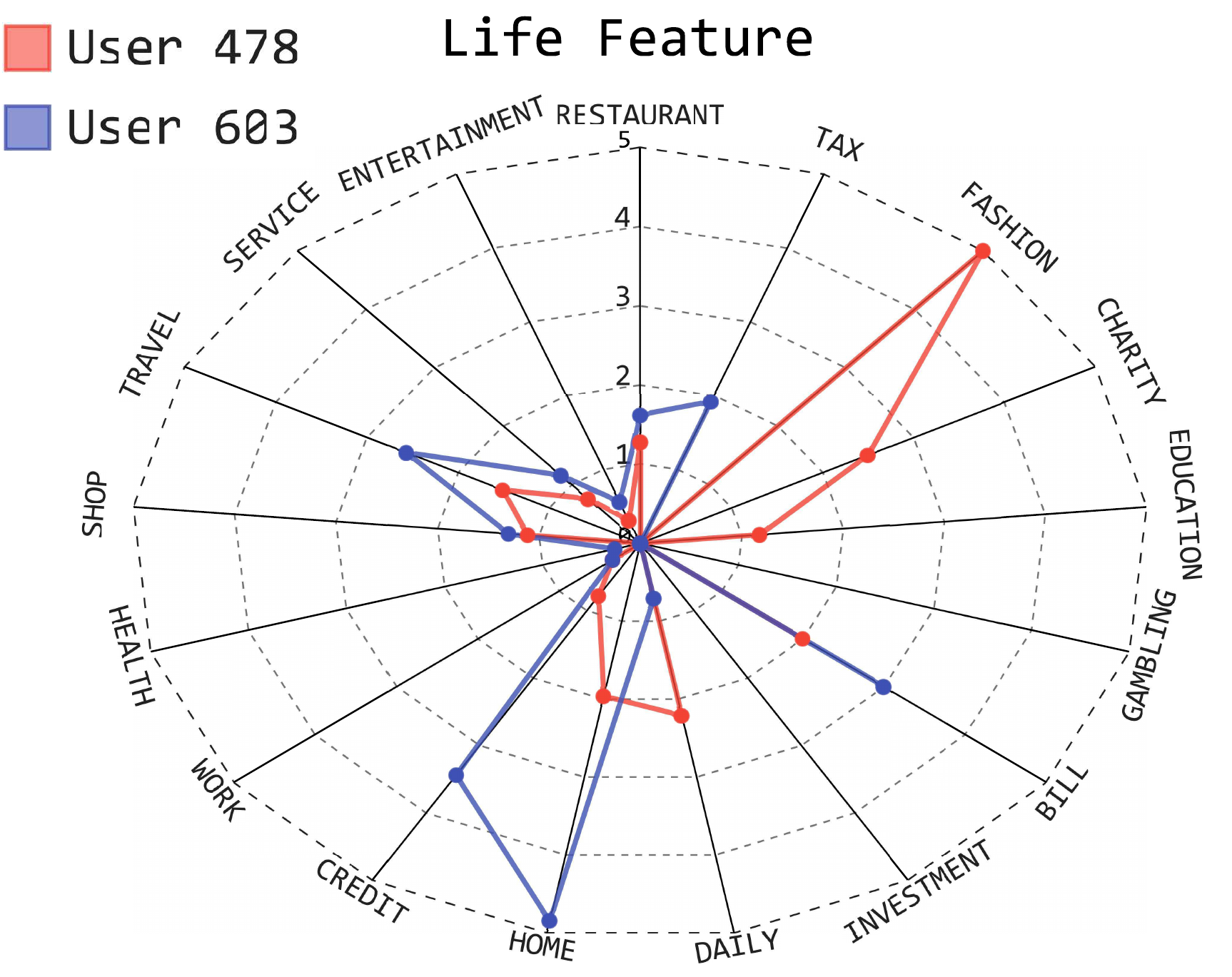}
\label{tab:lifesample}
\caption{Life features of user 604 and user 478.}
\end{figure}
As an example, Fig.~6 describes the life features of two arbitrary users: user 604 and user 478.
It shows that user 478 has a higher fashion score while user 603 has a higher home score. Therefore, a company should recommend clothes to user 478 and home items to user 603.

To show the pattern clearly, we negative the AIO Standard Deviations (SD) of the four groups in Fig.~4(b)-(e).
The AIOs negative SD curves show identical patterns as the overlap rates: consumers of adjacent groups have smaller differences in their AIO differences; the AIO similarity tends to decrease and then increase from group 1 to group 10.
Further, we draw the following conclusions from the above observations:
    1) Life features indicate the structures of consumers' purchase types;
    2) Life features reflect the purchasing choice characteristics, overlapping choices among different groups, and shopping scopes of consumers.

\section{Methodology}\label{sec:methodology}

\subsection{Framework Overview}

\begin{figure*}[!h] 
\includegraphics[width=\linewidth]{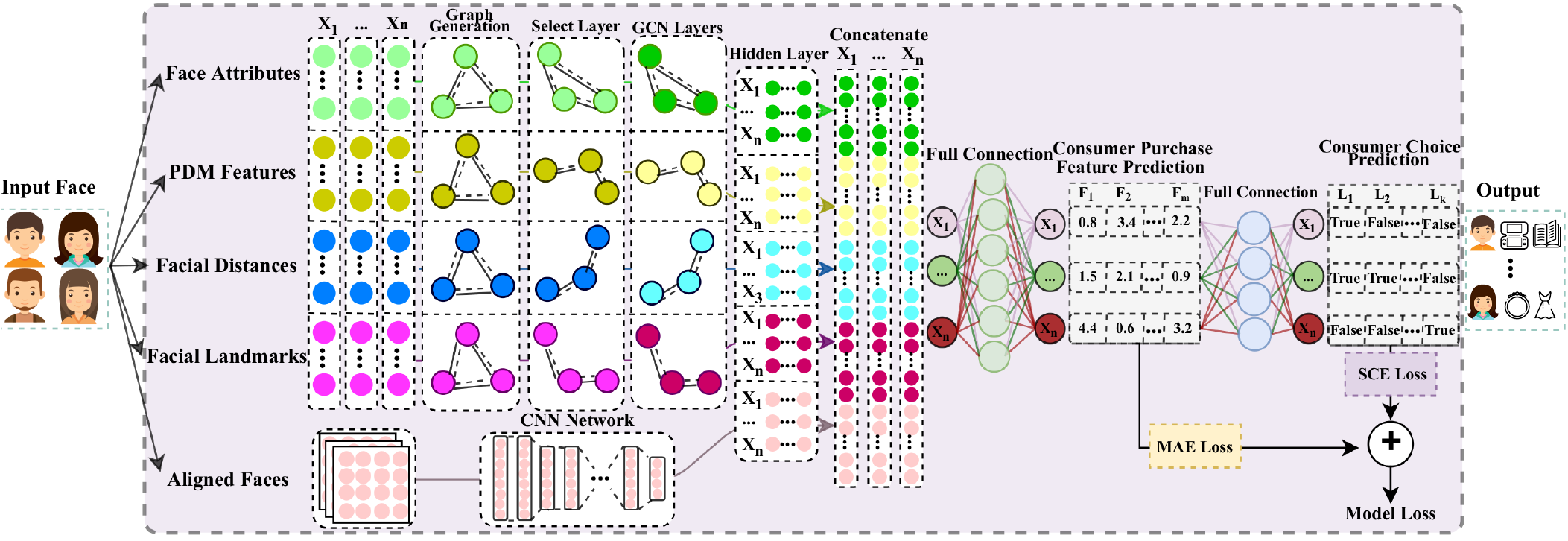}
\caption{The structure of our proposed model. $X_i$, $F_i$, $L_i$, MAE, and SCE denote a sample, consumer purchase features of the sample, purchasing choice labels of the sample, mean absolute error, and softmax cross entropy. '$\oplus$' denotes the sum of the MAE loss and the SCE loss.}
\label{model}
\end{figure*}

Our proposed framework predicts consumer behaviors with multi-model structured personal traits embedding.
Our model ({Fig.~\ref{model}}) consists of four steps, given a face dataset of $n$ face samples \{$X_1$,$X_2$,$X_3$,...,$X_n$\}:
(1) {Feature Engineering}---We design consumers' purchase features by considering their shopping capacity and compositions to label the data, and extract face descriptors of consumers by training a separate model for each sample $X$.
The features contain five aspects: face attributes (FA), point distribution model (PDM) parameters, facial distances (FD), facial landmarks (FL), and aligned faces (AF).
(2) {Selective GCN}---We organize the one-dimensional face descriptors of each sample into a weighted graph structure $g$ to generate multi-view face embedding $y$ with selected graph information.
(3) {CNN with Aligned Faces}---We extract the semanticity $sem$ from Aligned Faces (AF) by subsampling the AF and enhance the convolution ability of networks through Light Inceptions.
(4) {Hierarchical Labels and Prediction}---We combine the embedding information from both GCN and CNN to predict $m$ consumer purchase features \{$F_1$,$F_2$,$F_3$,...,$F_m$\} and thereby forecasting $k$ purchasing choices \{$L_1$,$L_2$,$L_3$,...,$L_k$\} for each consumer.

Specifically, we extract the face attributes, point distribution model (PDM) Parameters, facial landmarks, and aligned faces based on the work of~\cite{agustsson2017apparent,baltrusaitis2018openface,baltruvsaitis2014automatic,zadeh2017convolutional,baltruvsaitis2016openface,zhangposition}, and we propose to use GCN Units (a GCN layer with a Radial Basis Function (RBF) kernel~\cite{ravale2015feature}) to calculate the distances of consumer face features and subsequently the similarity connections. The similarities can be seen as edge weights and used for generating the graph structure of facial features. We also use a light CNN to analyze the hidden semantic information in the aligned faces. With the face information data and structures, we design a combined semi-supervised Selective Convolutional Network (SCN) model, which uses the images and the extracted information to predict consumer purchase features and shopping decisions, based on the work of~\cite{kipf2016semi,szegedy2015going}.

\subsection{Feature Engineering}\label{featureExtract}

\subsubsection{Purchase Feature Designation}

We define two categories of purchase features to reflect consumers' shopping capacity and compositions, in light of our discussion in Section~\ref{sec:features}.

\paragraph{Stratum Features.}
We define three types of stratum features of consumers, referring to Maslow's hierarchy of needs~\cite{mcleod2007maslow}:
\textit{Basic},
\textit{Social}, and
\textit{Self}.
Similar to Maslow's theory, each type represents a higher-level need compared to its precedent. 
We thereby identify three categories of users with weak, ordinary, and strong purchasing capacities. 
The three user categories have larger expense allocations on the higher-level needs from the first to the last. 

\paragraph{Life Features.}
We define consumers' life features by analyzing their purchasing choices and frequencies in 17 categories, borrowed from AIO (Activities, Interests, and Opinions) dimensions~\cite{boote1980psychographics} used for segmenting a target market: \textit{Restaurant}, \textit{Entertainment}, \textit{Service}, \textit{Travel}, \textit{Shop}, \textit{Health}, \textit{Work}, \textit{Credit}, \textit{Home}, \textit{Daily}, \textit{Investment}, \textit{Bill}, \textit{Gambling}, \textit{Education},  \textit{Charity}, \textit{Fashion}, \textit{Tax}. We will detail the labeling methods in Section \ref{sec:Label and Prediction}.

\subsubsection{Face Descriptor Extraction}
The prediction labels and the input data are extracted from the data set. The face information extraction is based on related research.
Previous studies~\cite{joo2015automated,zhang2015learning,al2014face,vernon2014modeling, baltruvsaitis2014automatic} show that valuable information such as nameable attributes, facial features, and landmark locations can be extracted from the face images and it can deliver better results when combining such information with raw images for training.
The work in~\cite{baltruvsaitis2014automatic} introduces a point distribution model (PDM) which can describe the face shape well. Therefore, besides extracting the PDM data as inputs, we obtain prominent face attributes, PDM parameters, and aligned faces using pre-trained models and then analyze landmark locations and facial feature distances based on the aligned faces. 
Fig.~8(a) shows the face information extraction flow. 

We apply a position-squeeze and excitation network (PSE)~\cite{zhangposition} to the LWFA dataset\footnote{http://vis-www.cs.umass.edu/lfw/} and the CelebA dataset\footnote{http://mmlab.ie.cuhk.edu.hk/projects/CelebA.html} to extract 16 facial attributes such as eyeglasses and attractiveness, and PSE achieves an average accuracy over 90\% on CelebA and over 85\% on LWFA. The aligned faces and landmark locations are from an open-source facial behavior analysis toolkit named OpenFace~\cite{baltrusaitis2018openface} with datasets: CMU Multi-PIE\footnote{http://www.cs.cmu.edu/afs/cs/project/PIE/MultiPie/Multi-Pie/Home.html}; HELEN and LFPW training subsets in 300W\footnote{https://ibug.doc.ic.ac.uk/resources/facial-point-annotations/}. The parameters of the point distribution model~\cite{baltruvsaitis2014automatic} are also extracted through OpenFace. With the landmark locations, we calculate 170 landmark distances as features such as forehead width.

\paragraph{Face Attributes.} Prominent face features include eyeglasses and attractiveness. We pre-train a position-squeeze and excitation network (PSE)~\cite{zhangposition} to extract 16 facial attributes and achieve the average accuracy of over 90\% and over 85\% on the CelebA~\cite{link2} and LWFA datasets~\cite{link1}, respectively.

\paragraph{PDM Features.} Point distribution model (PDM)~\cite{baltruvsaitis2014automatic} is a linear model that describes the rigid (location, scale, and rotation) and non-rigid face shapes. We use pre-trained models to generate PDM features and to obtain the shape similarity between faces.

\paragraph{Aligned Faces.} We obtain aligned faces from an open-source facial behavior analysis toolkit named OpenFace~\cite{baltrusaitis2018openface} with several datasets: CMU Multi-PIE~\cite{link3}; HELEN and LFPW training subsets in 300W~\cite{link4}.
We then calculate the normalized contour locations and distances based on aligned faces, which provide clear, centered face pictures with pure face semanticity information.

\paragraph{Facial Landmarks.} We obtain key similarity information about face contour and shape from landmark locations detected via shape models or CNN. We record $n$ landmarks \{$l_1$,$l_2$,$l_3$...,$l_n$\} by keeping their axes in the form of \{($x_1$,$y_1$),($x_2$,$y_2$),($x_3$,$y_3$),...,($x_n$,$y_n$)\}.

\paragraph{Facial Distances.} It reflects features of key components of faces such as mouth width and nose length. We design 170 facial distances to discover the similarity between faces in multiple aspects such as the general shape, key shape and contour, and face features.
We analyze the face from prominent attributes, general shape similarity, semanticity information.
We thereby embed faces from a more comprehensive sight than only using face images or a single type of descriptors. 
\begin{figure*} 
\center
\includegraphics[scale=1]{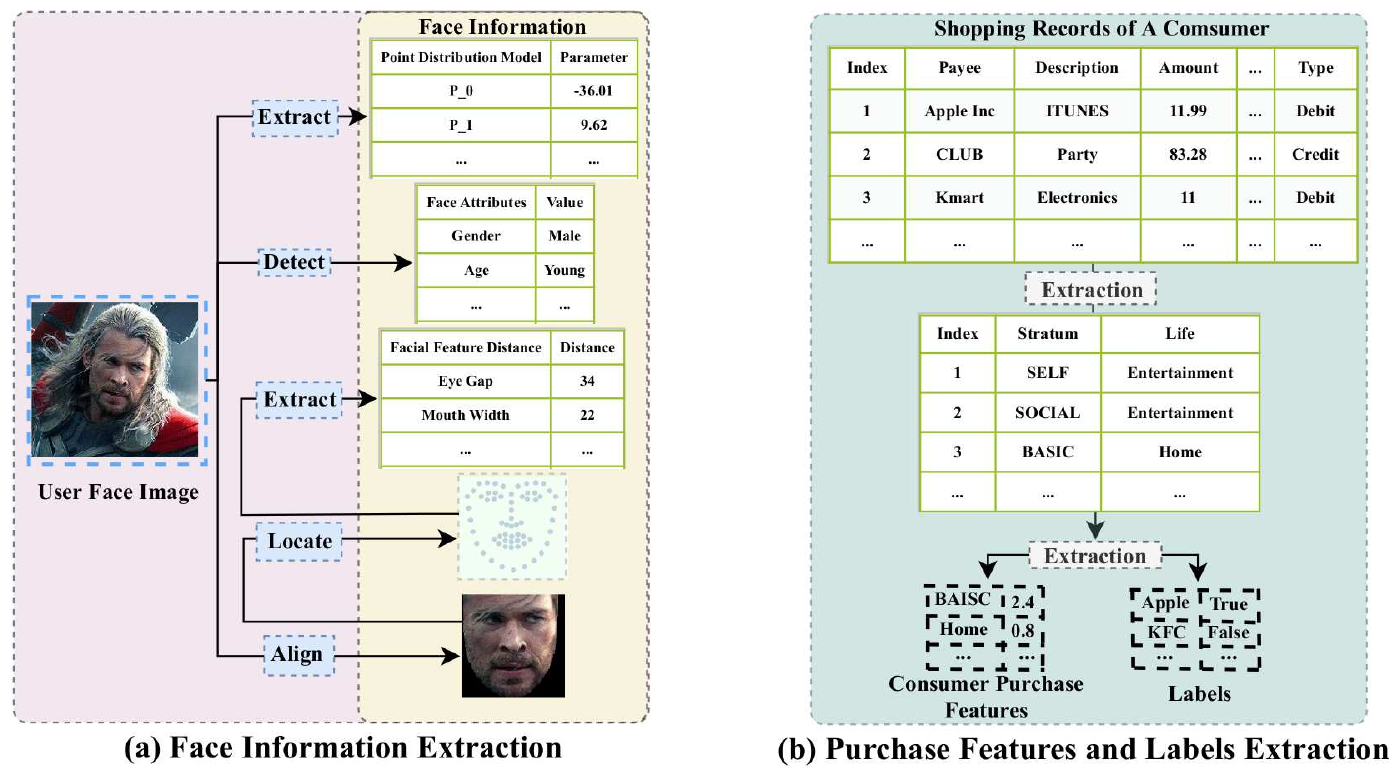}
\label{face_extract}
\caption{(a) Procedure for extracting face information; (b) Procedure for extracting consumer purchasing features and purchasing choice labels.}
\end{figure*}
%

\subsection{Selective Graph Convolutional Network}\label{2}
Our proposed SCN consists of two parts: Graph Convolutional Network (GCN) and Convolutional Neural Network (CNN). We design GCN by adapting the model in~\cite{kipf2016semi} to process the face information with the graph structure that we have already generated. Similarly, we adapt CNN from Inception~\cite{szegedy2015going} to process the aligned face images for extracting the semantic meanings. We combine five different aspects of face data to make our model more comprehensive.
The selective GCN generates elite graph embedding from original graph structures via {graph generation}, {graph selection}, and {graph convolution}.
\begin{figure*} 
\includegraphics[width=\textwidth]{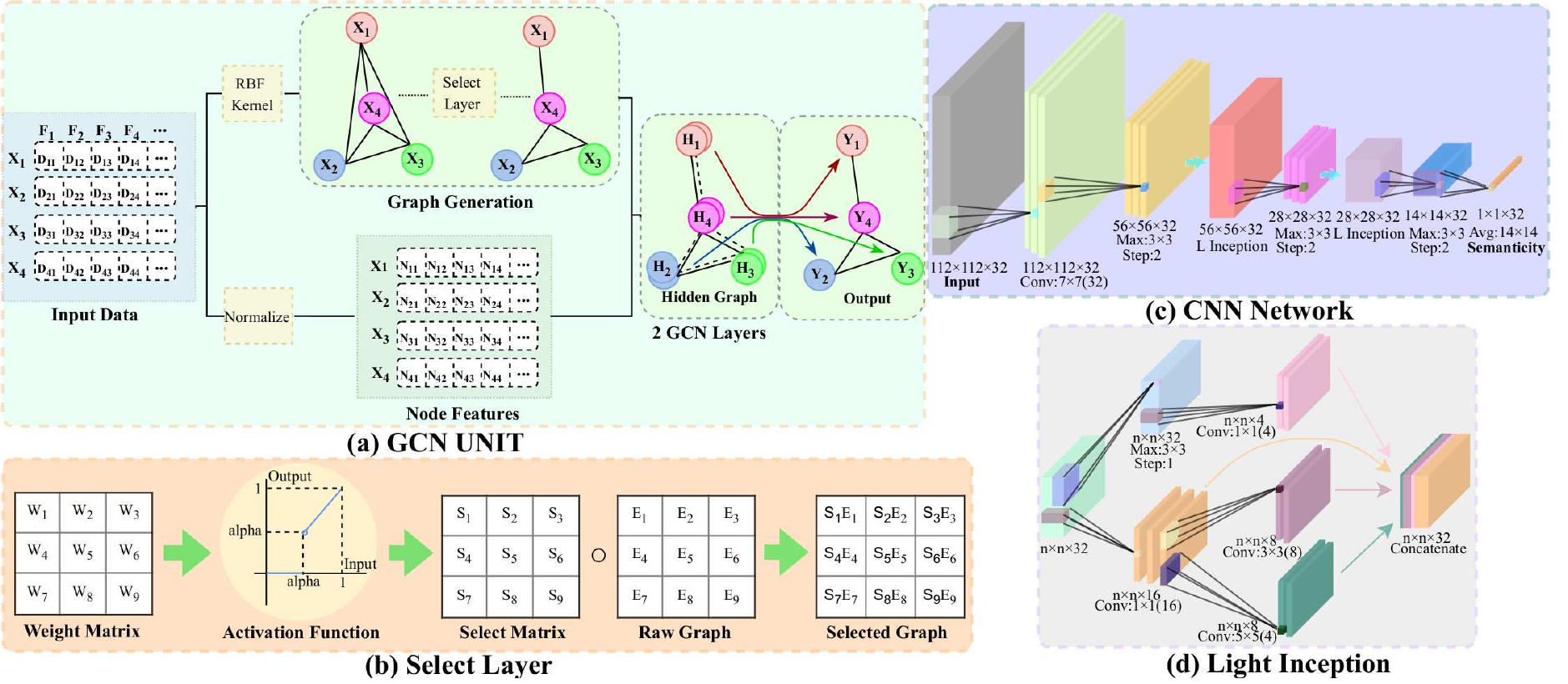}
\label{structure_details}
\caption{(a) Structure of a GCN unit. $X_i$, $H_i$, and $Y_i$ denote the raw node, hidden node, and the output node. $F_i$, $D_i$, and $N_i$ denote the feature dimension, feature data of nodes, and the normalized node feature data. (b) Procedure for the Select Layer. $W_i$, $S_i$, $E_i$, $S_iE_j$ denote the initial weight, the select weight, the raw edge weight, and the selected graph edge weight. The blue line in the Activation Function is the output curve, and the alpha in the axis is the threshold to drop edges. $\circ$ is the Hadamard product. (c) Structure of the semantic CNN network. (d) Structure of the light inception unit. For both (c) and (d), the first line of the description under each layer is the size of the outputs, and the second line is the operation before the layer, following by the kernel size and the kernel number. The operation of 'Max', 'Conv', 'Avg', and 'L Inception' denotes max-pooling, convolution, average-pooling, and Light Inception.} 
\end{figure*}

\subsubsection{Graph Structure Generation}
The original face information does not possess a graph structure. Inspired by the existing face clustering research~\cite{zhang2015learning}, we extract cluster information from faces as the additional input.
To this end, we use the RBF kernel to generate graph edges, based on face similarities, to represent the propagation abilities and to enhance our model. This way, we get a fully-connected graph $g$ of facial features based on cluster similarity.
The RBF kernel has been widely used for clustering and classification~\cite{kuo2014kernel,ravale2015feature}.
It measures similarity as exponential to the square Euclidean distance between two vectors:
\begin{equation}
\mathcal{K}(a,b) = RBF(a,b) = exp(-\frac{{\left \| a-b \right \|}^{2}_{2}}{\gamma }),\ \ \ \ a,b\in X
\end{equation}
where $a$ and $b$ denote two samples, $\gamma$ is the number of features.

\subsubsection{Graph Structure Selection}
We use a Select Layer to process the generated graph structure to reduce noises and useless edges before passing face information into GCN layers. Specifically, there exist many low weight edges indicating useless and false information in the graph structure we have generated, and it could be time-consuming if we pass the entire graph structure into the GCN. Therefore, we design the select layer to let the network drop the useless edges through training. {Fig.~9(b)} shows the data flow and the mathematical operations of the select layer. 
Specifically, we initial a weight matrix for the edges in the raw fully-connected graph and then feed the matrix to the activation function.
\begin{equation}
w_{i,j} = g \circ SelectLayer(Weight_{i,j}),
\end{equation}
\begin{equation}
SelectLayer(r)=\left\{
\begin{aligned}
0,\ \ \ \  & r\leq \alpha \\
r,\ \ \ \  & r>\alpha
\end{aligned}
\right.\ \ \ \  r \in \mathbb{R}
\end{equation}
where $g$ is the raw graph structure of face features generated in the previous step; $\alpha$ is the drop threshold; $Weight_{i,j}$ is a cell of the weight matrix of $g$, indicated by the index of samples $i$ and the index of features $j$.
The activation function will drop the edge if the edge weight is lower than the threshold. Finally, we consider the output matrix of the activation layer as the select matrix to optimize the propagation of raw graph---we calculate the Hadamard product of the select matrix and the raw graph as the selected graph $w$.

\subsubsection{GCN Layer}.
The facial feature information usually provides the hidden similarity information to help the models realize and distinguish different types of faces. Therefore, we apply the GCN Unit shown in Fig.9(a) to analyze the extracted facial feature information.

Given the normalized feature data and the purified graph structure, we further analyze the relationship between different faces by analyzing graph structures and face features using GCN. Basically, GCN is an efficient spatial-based graph neural network that uses the neighbour nodes to represent the raw nodes and does not need to calculate the eigenvectors and eigenvalues. Here, we predict node labels through propagation rules using a multi-layer GCN~\cite{kipf2016semi}:
\begin{equation}
H^{(l+1)}= \widetilde{D}^{-\frac{1}{2}}\widetilde{A}\widetilde{D}^{-\frac{1}{2}}H^{(l)}W^{(l)}
\end{equation}
where $H^{(l)}$ is the $l$th GCN hidden layer; $W^{(l)}$ denotes the weight matrix of the $l$ layer; $\widetilde{A} = A + I_{N}$ is the processed adjacency matrix from the select layer after adding a identity matrix $I_{N}$; $\widetilde{D}$ is a degree matrix of $\widetilde{A}$. We use two GCN layers in each unit and take the output of the activated hidden layer as face feature hidden information.
\begin{equation}
h = GCN(w)*normalize(X)
\end{equation}
\begin{equation}
y = Relu\big(LeakyRelu(h)\big)
\end{equation}
where $h$ denotes the hidden layer of GCN; Leaky Relu and Relu are used to help the model gain better graph representations; $y$ is the embedding of face information for FA, PDM, FD, or FL.

\subsection{Convolutional Neural Network with Aligned Faces}
We adopt CNN to extract the hidden semanticity in faces as the embedding information, in light of the outstanding performance of CNNs in analyzing pictures~\cite{szegedy2015going}.
Our CNN structure processes the aligned faces following a three-step procedure: subsample, Light Inception, and semanticity.

\subsubsection{Subsample}
We design the CNN structure to find the balance point between effectiveness and efficiency. Therefore, we apply the subsampling method to construct our network. First, we only apply the 7 $\times$ 7 convolution layer at the beginning of the network because a larger-size layer can catch a wider range of information and provide a better environment for the following layers. However, the large scale layer will be very time-consuming, so we only apply the kernel layer at the bottom of the network. Second, we adopt the 3 $\times$ 3 max-pooling kernel in step two to help reduce the scale of images. The pooling layer with kernel size three could smoothly decrease the data scale with a moderate computation burden. Therefore, the subsample structure should be:
\begin{equation}
subsample = Max_{3 \times 3}\big(Conv_{7 \times 7}(layer_l)\big)
\end{equation}
where $layer_l$ is the current layer; $Max_{3 \times 3}$ is a max-pooling layer with a 3 $\times$ 3 kernel; $Conv_{7 \times 7}$ is a 7 $\times$ 7 convolution layer.

\subsubsection{Light Inception}
Some information may lose during the downsampling of faces. Therefore, we apply the convolution layer to extend the data range and information. Specifically, we design the Light Inception showing the overall structure based on the method of Inception~\cite{szegedy2015going} to extract deeper hidden information efficiently. The Inception layer is a 3-depth unified structure aimed to extract deep information effectively.
The idea is to use a large-size convolution layer after the small-size kernel layer and to use a convolution layer after the max-pooling layer. Different from the conventional Inception, we keep the same 1 $\times$ 1 kernels before high-level convolution layers.
We define the large-size convolution and max-pooling convolution extension as follows:
\begin{equation}
I_p = Conv_{1\times1}\big(Max_{1\times1}(layer_l)\big)
\end{equation}
\begin{equation}
I_{Con_{3\times3}} = Conv_{3\times3}\big(Conv_{1\times1}(layer_l)\big)
\end{equation}
\begin{equation}
I_{Con_{5\times5}} = Conv_{5\times5}\big(Conv_{1\times1}(layer_l)\big)
\end{equation}
where $layer_l$ is the current layer; $I_p$, $I_{Con_3\times3}$, and $I_{Con_5\times5}(layer_l)$ are the Inception max-pooling layer, the Inception convolution layer with a $3 \times 3$ kernel and a $5 \times 5$ kernel; \textit{Conv} is the normal convolution layer; both the max-pooling and convolution layers use step one. To get the most of CNN, we use 1 $\times$ 1 convolution layers before the 3 $\times$ 3 and 5 $\times$ 5 convolution layers and add the extension information to the original 1 $\times$ 1 convolution:
\begin{equation}
LightInception = F_c\big(I_p(con),I_{Con_{3\times3}}(con),I_{Con_{5\times5}}(con),con\big)
\end{equation}
\begin{equation}
con = Con_{1\times1}(layer_l)
\end{equation}
where $layer_l$ is the previous layer of the Light Inception Layer and $con$ is the public $Con_{1\times1}$ layer for further convolution. We merge the different convolution output as the Light Inception output.

\subsubsection{Semanticity}
We combine subsample and Light Inception structures to analyze faces effectively and efficiently---We apply the subsample-LightInception (SLI) twice to construct the extraction part of our network and adopt an average-pooling layer to accelerate convergence.
Considering Light Inception as a convolution layer, we do not use additional convolution layers in the second SLI but adopt an average-pooling layer to accelerate convergence.
\begin{equation}
sem = Avg_{14 \times 14}\Big(SLI\big(SLI(input)\big)\Big)
\end{equation}
where $input$ is the aligned consumer faces; $Avg_{14 \times 14}$ denote the average-pooling layer with the kernel 14 $\times$ 14; and $sem$ means the semanticity embedding information in faces.
Specifically, we use 1 $\times$ 1 convolution layers before 3 $\times$ 3 and 5 $\times$ 5 convolution layers. Then, we concatenate four different types of convolution information and apply a batch normalization layer~\cite{ioffe2015batch} to process the concatenated information.

\subsection{Hierarchical Labels and Prediction}\label{sec:Label and Prediction}
We extract the hierarchical labels (indicating whether a consumer purchased at a company) based on statistics of consumers' purchasing information, as shown in Fig.~8(b). 
We label each transaction record according to amount, category and description following the life feature and stratum feature structure. Then, we calculate the ratio scores of different needs and the frequencies of shopping in each life aspect, respectively.
We extract consumers' purchasing features and purchasing choices as follows:
\begin{equation}
F_{i}(life_y) = \frac{5*frequency_i(life_y)}{\max\limits_{x\in[1,n]} frequency_x(life_y)}, i \in [1,n]
\end{equation}
\begin{equation}
F_{i}(stratum_y) = \frac{5*expense_i(stratum_y)}{\sum_{x\in[1,3]}{expense_i(stratum_x)}}, i \in [1,n]
\end{equation}
\begin{equation}
L_j=\left\{
\begin{aligned}
0 & , & L_j \in history \\
1 & , & L_j \notin history
\end{aligned}
\right.\ \ \ \ j \in [0,k]
\end{equation}
where \textit{frequency} and \textit{expense} denote the times and total spending of a consumer' purchases spend on a specific feature; $F_{i}(feature_y)$ is the score of the $i$th consumer on the $y$th feature. Since {life features} focus on the frequency ranks of consumers, we compress the life shopping frequency to the range of 0 to 5 according to the peer maximum. Similarly, we calculate and compress the shopping ratio of each type according to the expense sum, given that {stratum features} concentrate on the ratio of each category of consumers. {purchasing choice labels} are determined by the consumer shopping history.

Our proposed model combines the hidden information from different views, four types of low features, and the aligned faces.
With the extracted face information, we predict consumers' purchasing features and purchasing choice labels with a full-connection layer. In particular, we concatenate the five-aspect information to obtain the relationship between face and behavioral traits and then use behavioral traits $F$ with $softmax$ to predict purchasing choices $L$:
\begin{equation}
F = W_F*concatenate(y_{FA},y_{PDM},y_{FD},y_{FL},sem)+b_F
\end{equation}
\begin{equation}
L = softmax(W_L*F+b_L)
\end{equation}
where $y_{FA}$, $y_{PDM}$, $y_{FD}$, $y_{FL}$, and $sem$ are the embedding on five aspects: FA, PDM, FD, FL, and AF; $W_F$, $W_L$, $b_F$ and $b_L$ are the weight matrix and the bias vector of purchasing features and purchasing choices.

We apply \textit{mean absolute error (MAE)} and \textit{softmax cross-entropy (SCE)} to measure the prediction of consumer purchase features and purchasing choices, respectively.
This leads to similar ranges of the loss functions for both prediction, thereby balancing the performance of both predictions. We define the overall loss function as follows:
\begin{equation}
Loss(F,{F}',L,{L}') = \frac{1}{n}\big(\alpha \cdot MAE(F,{F}') + \beta \cdot SCE(L,{L}')\big)
\end{equation}
\begin{equation}
MAE(F,{F}')=\sum _{i}\sum _{j}\frac{1}{m}\cdot \left | F_{i,j}-{F}_{i,j}' \right |
\end{equation}
\begin{equation}
SCE(L,{L}')=\sum _{i}\sum _{o}\frac{1}{k}\cdot  -{L}_{i,o}' *log(L_{i,o})
\end{equation} 
where ${F}'$ and ${L}'$ denote the ground truth; $F$ and $L$ are the prediction results;
$\alpha$ and $\beta$ are the hyper-parameters to control the weights of MAE and SCE, respectively;
$n$, $m$, and $k$ represent the number of samples, features, and labels, respectively; $i$, $j$, and $o$ denotes the index of samples, features, and labels, respectively.

\section{EXPERIMENTS}\label{sec:exp}

\subsection{Experimental Setting}
We studied a dataset from a local financial services company, which is described in Section~\ref{sec:features}.
We chose four companies with the most consumers to test our proposed model, i.e., SCN, through two sets of experiments.
We first compare the GCN unit with some traditional multi-label methods to test the effectiveness of the generated graphs; then, we compare different models and study the performance of this semi-supervised model under different conditions, especially the impact of adding consumer purchase features to the models.
Specifically, we first compare SCN with the traditional methods under various conditions to show the effectiveness of the structured behavioral traits embedding, the GCN unit, and the select layer. Then, we analyze the feature correlation to show the reliability of our model.
We use precision, recall, and F1 scores as the evaluation metrics to mitigate the impact of imbalanced distributions of positive and negative samples.
To test the effectiveness of Light Inception, we examine the models with and without image inputs. SCN uses all five types of data, and SCN (N) only uses the facial feature information data.

We directly use the multi-label package in~\cite{scikit-learn} to realize the baselines: extreme Gradient Boosting (XGBOOST)~\cite{chen2016xgboost}, Linear Logistic, C-Support Vector Classification (SVC) and Linear Support Vector Classification (Linear SVC).

\subsection{Effectiveness of Structured Behavioral Traits Embedding}
After proving the effectiveness of GCN structure, we test SCN under several conditions: the models with and without image inputs and different fractions of training data. Table~\ref{tab:Self-comparison} shows that both SCN and SCN(N) have better performance than the single GCN unit, showing that consumer purchase features can provide useful information to the models. The consumer purchase features have an acceptable range in MAE, i.e.,  0.4 to 0.5 in the range of 0 to 5. SCN has higher F1 and recall than models without images under all the conditions. This observation shows Light Inception can make the model more robust in handling disproportionate data. The SCN can still have good performance when using 8\% of the data for training.
We compare SCN and SCN(N) with a single GCN unit to demonstrate the impact of behavioral traits embedding.
{Table~\ref{tab:Self-comparison}} shows both SCN and SCN(N) have better performance than the other methods, indicating consumer purchasing features provide useful information to the models. The results have an acceptable MAE between 0.4 to 0.5 in the range of (0, 5). SCN has higher F1 and recall than SCN(N) almost in all circumstances, showing the Light Inception improves the model's robustness. SCN still achieved comparable performance when using less than 10\% data for training, revealing the effectiveness of semi-supervised learning.
\begin{table*}
  \caption{Comparison between models.Fraction denotes the fraction of labeled data. `(N)' indicate the input contain no images. MAE is the loss of middle features. Standard deviations are in parentheses. Traditional algorithms and GCN do not predict consumer purchase features.}
  \label{tab:Self-comparison}
   \center
   \resizebox{\textwidth}{!}{
  \begin{tabular}{cc|ccccc}
  \toprule
   & Fraction&Macro-precision & Macro-recall & Macro-F1 & Model Macro-F1 & MAE \\
   \midrule
    LINEAR SVC & all & 0.268(9.14e-2) & 0.5(0) & 0.342(7.23e-2) & 0.349 & /\\
   SVC & all & 0.307(6.70e-2) & 0.498(3.30e-3) & 0.377(4.98e-2) & 0.380 & /\\
   XGBOOST & all & 0.447(4.79e-2) & 0.482(2.36e-2) & 0.463(3.12e-2) & 0.464 & /\\
   LINEAR LOGISTIC & all & 0.515(4.34e-3) & 0.520(3.12e-2) & 0.517(3.72e-2) & 0.517 & /\\
   GCN & all & 0.593(1.62e-1) & 0.528(2.87e-2) & 0.551(7.41e-2) & 0.559 & / \\
   \midrule
   Semi-SCN(N) & 13\% & 0.550(8.22e-2) & 0.514(1.37e-2) & 0.530(3.78e-1) & 0.532 & 0.437\\
   Semi-SCN(N) & 35\% & 0.563(7.63e-2) & 0.526(2.23e-2) & 0.542(4.11e-1) & 0.544 & 0.438 \\
   SCN(N) & all & \textbf{0.626(2.14e-1)} & 0.534(5.71e-2) & 0.567(1.03e-1) & 0.577 & 0.498 \\
   Semi-SCN & 13\% & 0.549(8.33e-2) & 0.530(6.94e-2) & 0.539(7.47e-2) & 0.539 & 0.378 \\
   Semi-SCN & 35\% & 0.566(4.02e-2) & 0.550(4.41e-2) & 0.558(4.20e-2) & 0.558 & 0.426 \\
   SCN & all & 0.604(7.75e-2) & \textbf{0.558(2.70e-2)} & \textbf{0.578(3.68e-2)} & \textbf{0.580} & 0.501\\
\bottomrule
\end{tabular}}
\end{table*}
\begin{figure*}
\includegraphics[width=\textwidth,height=50mm]{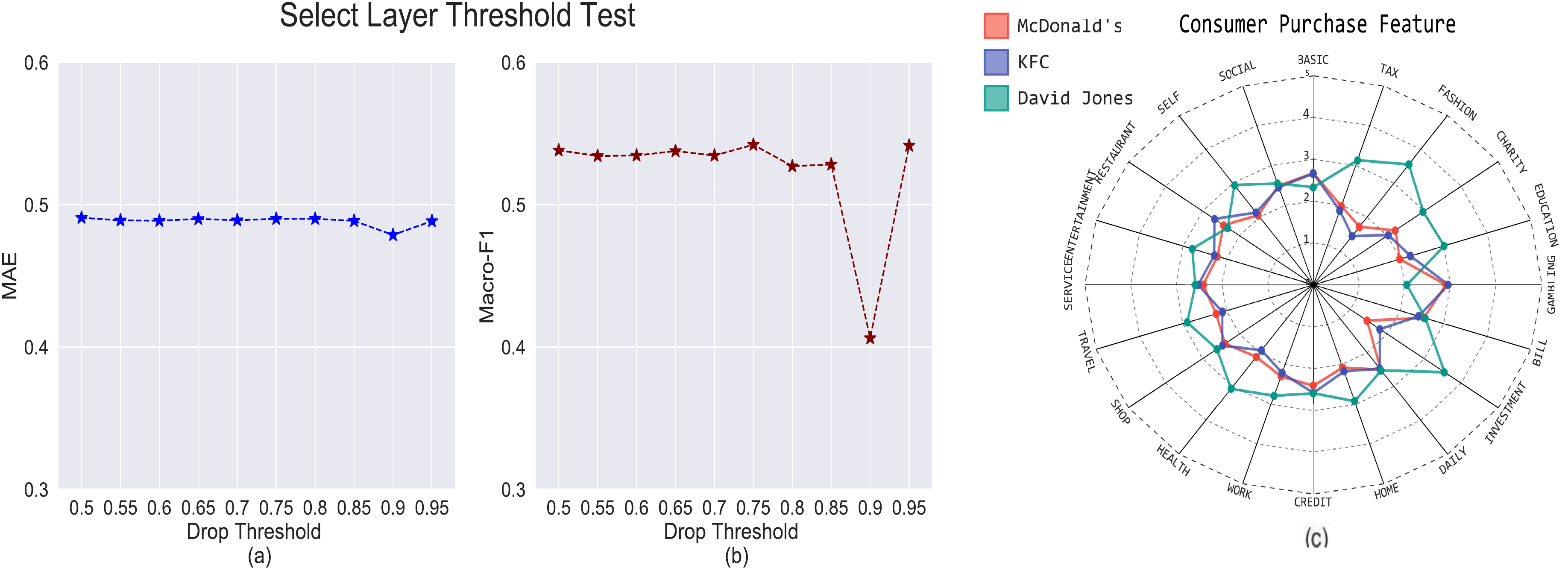}
\caption{(a): MAE, (b): Macro-F1 scores, under the different drop thresholds of the select layer. The weights of edges below the threshold are not shown. (c): Purchasing features of consumers of three companies.}
\label{ablation}
\end{figure*}

\subsection{Graph Effectiveness in the GCN Unit}
We compare the GCN unit with traditional methods to demonstrate the advantages of generating the graph structure of non-network data in image analysis. To ensure the fair comparison, we used only the extracted facial feature information as the input. {Table~\ref{tab:Self-comparison}} shows the GCN unit achieved a significant improvement in precision and F1 score over traditional algorithms: XGBOOST and Linear Logistic achieved the average performance while SVM-related algorithms performed worst. The results demonstrate the effectiveness of the graph structure.
\begin{figure*}[!hbt]
\centering
\begin{minipage}[t]{0.48\textwidth}
\centering
\includegraphics[width=\textwidth,height=66mm]{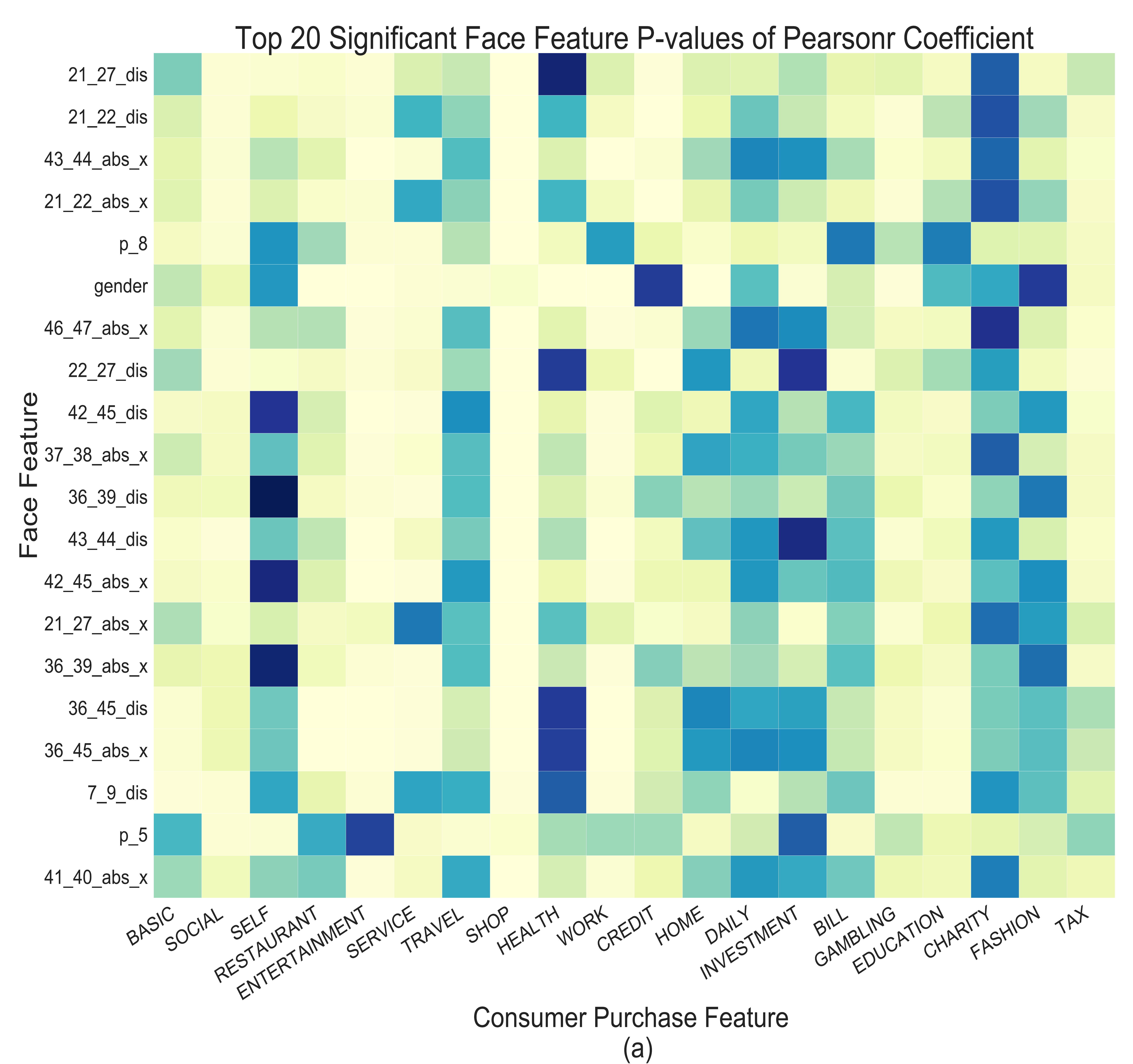}
\end{minipage}
\hspace{2mm}
\begin{minipage}[t]{0.48\textwidth}
\centering
\includegraphics[width=\textwidth,height=66mm]{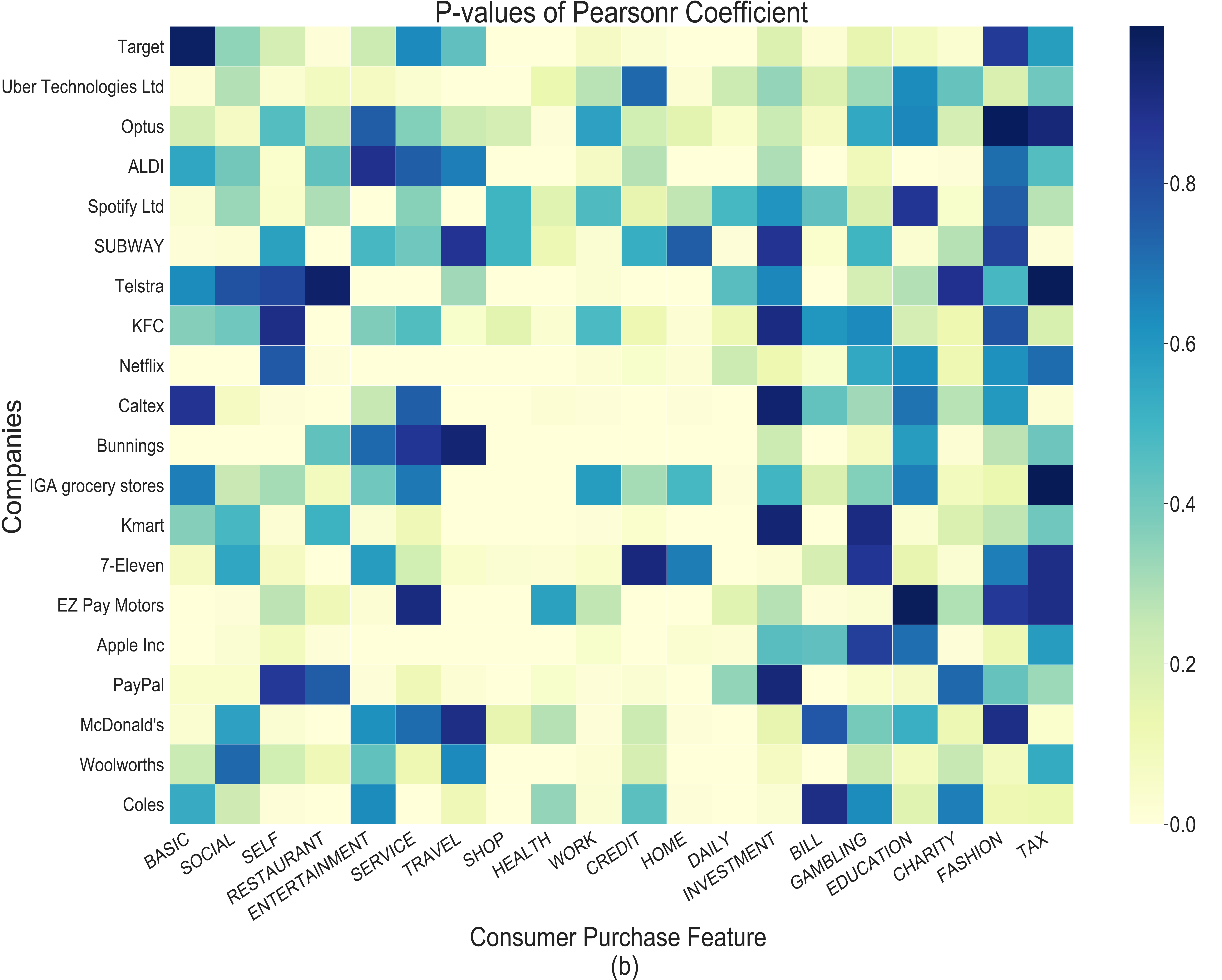}
\end{minipage}
\caption{(a): P-values of top 20 significant face features with consumer purchase features. (b): P-values of consumer purchase features with purchasing choices. Features with p as the first letter are PDM features; numbers means the landmark location; abs means absolute distance; dis denotes Euclidean distance.}
\label{facevalue}
\end{figure*}

\subsection{Effectiveness of the Select Layer}
We tested the impact of the select layer by varying the threshold from 0.5 to 0.95 at the increment of 0.05, where the weight matrix of the select layer is set randomly within the range of (0.5, 1).
Besides, we set the batch size to 64 and the early stop epoch to 50.
{Figs.~\ref{ablation}a-b} show the MAE and Macro-F1 Scores under different thresholds. We observed that the MAE kept stable while the Macro-F1 score peaked at 0.75 and dropped drastically at 0.9. The results indicate good performance despite the noises hidden in edges. Generally, the select layer can learn the graph structure along with dropping useless edges to reduce the computation time. Meanwhile, it can keep most of the information.

\subsection{Feature Correlation Analysis}

{Fig.~\ref{ablation}c} shows purchase features of consumers of three companies: {McDonald's}, {KFC} and {David Jones}. We normalized the scores and enlarge the mean scores to show clear differences. We observed consumers of KFC and {McDonald}'s both have high scores in \textit{Basic} yet low scores in \textit{Social} and \textit{Self}. This is reasonable because they have the same brand symbol: fast food.
Interestingly, their consumers gamble more frequently than the consumers of David Jones, a fashion company.
In comparison, consumers of David Jones generally have stronger shopping capacity; they purchase more frequently in more categories such as \textit{Self}, \textit{Fashion}, and \textit{Investment} yet less frequently in \textit{Basic}.
{Figs.~\ref{facevalue}a-b} show the Pearson correlation coefficient p-values~\cite{benesty2009pearson} of facial information to consumers' purchasing features and consumers' purchasing features to their purchasing choices, respectively. We found only 14 features (except \textit{self}, \textit{travel}, \textit{health}, \textit{daily}, \textit{investment}, and \textit{charity}) have significant correlations with the facial information. All the companies in {Fig.~\ref{facevalue}b} have strong correlations with multiple features.

\subsection{Case Examples}
\begin{figure*}[!t]
\center
\includegraphics[width=0.85\textwidth]{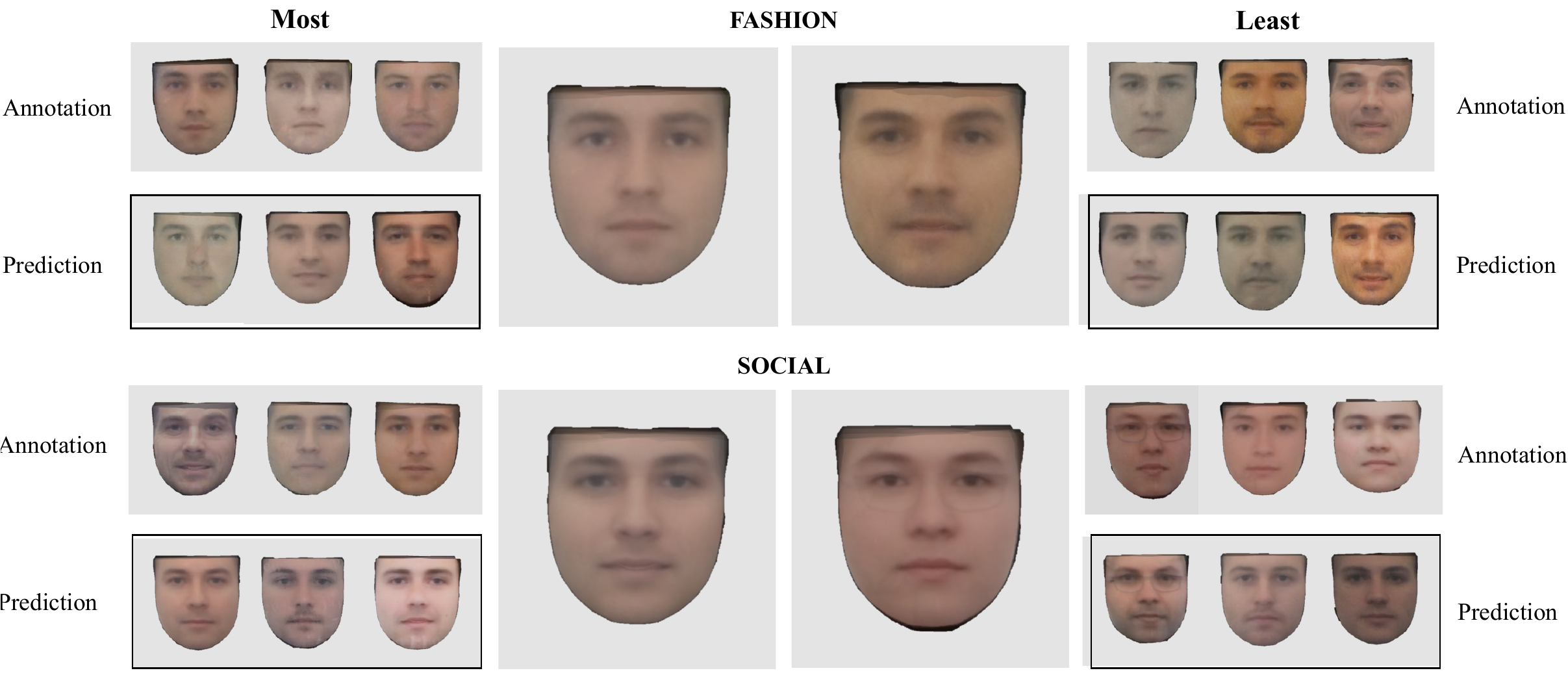}
\caption{Consumer faces indicative of strong and weak purchasing tendencies in two categories: \textit{Fashion} and \textit{Social}.
For both dimension: (middle) left and right are the average faces with overall high responses and low responses to the features; (from left to right) other faces represent an averaged example of a face set in the rating order from highest to lowest; the averaged examples contain (top-left) annotation and (bottom-left) prediction, marked with black rectangles.}
\label{average_face}
\end{figure*}

We use \textit{Fashion} and \textit{Social}, which belong to stratum features and life features, respectively, to demonstrate the face features indicative and non-indicative of consumer features in Fig.~\ref{average_face}.
We only show the generated average male faces to protect consumer privacy.

\begin{figure*}[!t]
\center
\includegraphics[width=0.95\textwidth,height=60mm]{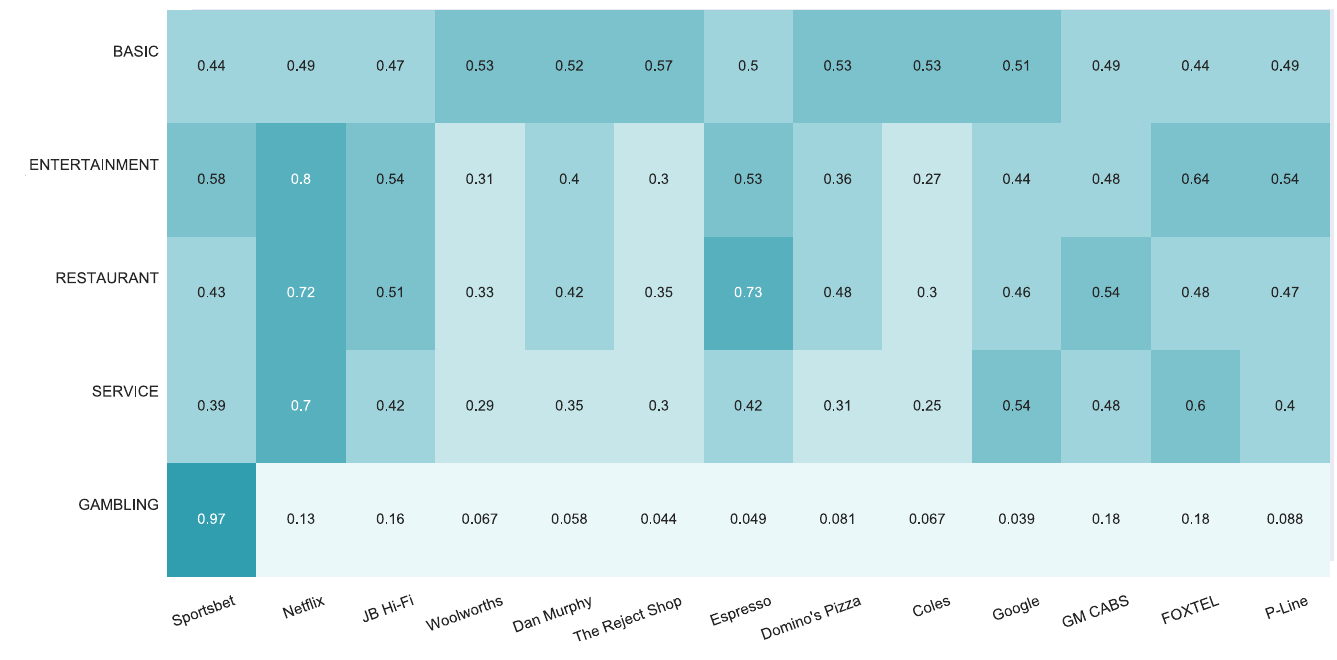}
\caption{Weighted purchasing features of consumer of different companies.}
\label{importance_analysis}
\end{figure*}

We further analyzed the consumer features of different companies.
First, we calculated the inverse shopping frequency $I_{s}$ as an additional weight to enlarge the patterns of low-frequency shopping companies.
Then, we took the sum of the weighted purchase features to represent the consumer characters of companies:
\begin{equation}
I_{Feature_{c,i}}=\frac{I_{s}(c)*\sum Feature_{c,i}}{SD(Feature_{c,i})},\ \ \ \ I_{s}(c)=\log\frac{N}{|c|}
\end{equation}
where $I_{Feature_{i}}$ is the weighted score of company $c$ on $Feature_{i}$; $SD$ is the standard deviation.
Fig.~\ref{importance_analysis} shows some examples of results, where we normalize stratum and life features separately to the range of $[0,1]$.
The results show both \textit{Sportsbet} and \textit{Netflix} had high scores in \textit{Gambling} and \textit{Entertainment}.
Interestingly, for a customer of \textit{JB Hi-Fi}, \textit{Dan Murphy} or \textit{Woolworths}, if he frequently purchases alcoholic and electronic products, then he tends to spend more on \textit{Entertainment} but less on \textit{Basic} (daily consumption) products.

\section{Related Work}

Face information extraction has been a research topic for a long time.
Traditionally, researchers focus on extracting low-level information such as face attribute information~\cite{zhangposition,ren2017time}, gaze information, face landmark information, emotion state~\cite{baltrusaitis2018openface} from face images, or extracting embedding information of faces for voice matching~\cite{horiguchi2018face} and face verification~\cite{wang2017normface,li2017integrated}.
Recently, researchers begin to transform the facial information to infer the higher-level information such as personality traits theory: the BIG 5~\cite{mccrae1992introduction}, the BIG 2~\cite{abele2007agency} and the 16 PF~\cite{qin2018modern}.
The high-level information extracted from human faces is widely used in the election prediction~\cite{joo2015automated} and social relationship prediction~\cite{zhang2015learning}. Furthermore, researchers also use the personality traits theory to simulate job interview results based on images~\cite{guccluturk2017multimodal,escalante2018explaining}.
Meanwhile, predictions of purchase behaviours have emerge as an application and extension of traditional recommendation research to harness various sources of heuristic information in e-commerce~\cite{yao2018collaborative}. A primary limitation of the personality traits theory is that it does not fit the purchasing behavior well~\cite{jacoby1969personality,lastovicka1988improving,aaker1997dimensions}, considering that personality traits do not contain social and financial information~\cite{punj1983interaction,allsopp1986distribution,foxall1988personality}.
Therefore, we propose and believe that it is necessary to use consumer purchase features to characterize consumers through behavioral traits from a business perspective other than from a personality perspective.

We develop our work based on the previous efforts in~\cite{kipf2016semi,szegedy2015going,joo2015automated,zhang2015learning}, yet we transfer the election prediction to the more complicated purchase behavior prediction and use data-driven features and model structures to extract hidden information from the faces. On the contrary, the method in~\cite{joo2015automated} does not consider the data characteristics and uses a straightforward approach.
The work of~\cite{zhang2015learning} demonstrates the effectiveness of cluster information of human facial features in improving the performance of DCN. However, such a model pre-assumes the cluster number.
Instead, we consider the similarity between faces as edges and propose to transform the cluster information into the similarity between nodes: the more similarity, the more effects. Therefore, we use the GCN model, a widely used model for the prediction of graph data~\cite{kipf2016semi}, to process the graph structure.
Besides, we apply the CNN structure in our model as the additional information to find the critical edges and to propagate the labels, given its ability to extract the hidden information from images~\cite{zhang2015learning}.

\section{Conclusion}

In this paper, we propose a hierarchical model that combines face embedding with structured behavioral traits embedding and demonstrate its better performance in predicting consumer purchases.
Face embedding contains face information from multiple aspects such as shape, contour, and semanticity, while structured behavioral traits capture consumer behaviors such as financial stratum and shopping tendencies.
We design a selective graph convolution unit to process the data without a graph structure and to provide additional structural similarity information to enhance the model. We also adopt an Inception structure to make a light version of the model.
A significant advantage of our model is to exploit human faces for hidden clues about consumers' personal traits and shopping behaviors. It can serve as a promising solution to cold-start scenarios, which lack consumers' behavior information.

\bibliographystyle{unsrt}
\bibliography{refs}

\end{document}